\documentclass[letterpaper, 10 pt, conference]{ieeeconf}  
\IEEEoverridecommandlockouts                              
\usepackage[pangram]{blindtext}
\usepackage{comment}
\usepackage{amsmath,amsfonts,amssymb}
\let\labelindent\relax
\usepackage{prettyref}
\usepackage{mathrsfs}
\usepackage{graphicx}
\usepackage{wrapfig}
\usepackage{subfig}
\usepackage{MnSymbol}
\usepackage{multirow}
\usepackage{booktabs}
\usepackage{algorithm}
\usepackage[table]{xcolor}
\usepackage[noend]{algpseudocode}
\usepackage{highlight}
\usepackage{enumitem}
\usepackage{hhline}
\usepackage
[backend=bibtex,
bibstyle=ieee,
citestyle=numeric,
mincitenames=1,
maxcitenames=2,
natbib=true,
doi=false,
isbn=false,
url=false,
eprint=false]{biblatex}
\usepackage[colorlinks,allcolors=gray,hypertexnames=true]{hyperref}

\algnewcommand{\LineComment}[1]{\State \(\triangleright\) #1}
\algdef{SE}[DOWHILE]{Do}{doWhile}{\algorithmicdo}[1]{\algorithmicwhile\ #1}
\newcommand*{\colorboxed}{}
\def\colorboxed#1#{%
  \colorboxedAux{#1}%
}
\newcommand*{\colorboxedAux}[3]{%
  \begingroup
    \colorlet{cb@saved}{.}%
    \color#1{#2}%
    \boxed{%
      \color{cb@saved}%
      #3%
    }%
  \endgroup
}

\newrefformat{fig}{Figure~\ref{#1}}
\newrefformat{par}{Section~\ref{#1}}
\newrefformat{appen}{Appendix~\ref{#1}}
\newrefformat{sec}{Section~\ref{#1}}
\newrefformat{sub}{Section~\ref{#1}}
\newrefformat{table}{Table~\ref{#1}}
\newrefformat{ass}{Assumption~\ref{#1}}
\newrefformat{alg}{Algorithm~\ref{#1}}
\newrefformat{def}{Definition~\ref{#1}}
\newrefformat{thm}{Theorem~\ref{#1}}
\newrefformat{lem}{Lemma~\ref{#1}}
\newrefformat{step}{Step~\ref{#1}}
\newrefformat{ln}{Line~\ref{#1}}
\newrefformat{eq}{Equation~\ref{#1}}
\newrefformat{pb}{Problem~\ref{#1}}
\newrefformat{it}{Item~\ref{#1}}
\newrefformat{te}{Term~\ref{#1}}
\def\Eqref Eq:#1:{\eqref{eq:#1}}
\newrefformat{Eq}{Equation~\Eqref#1:}

\newcommand{\TE}[1]{\textbf{#1}}

\newcommand{\argmin}[1]{\underset{#1}{\text{argmin}}}
\newcommand{\argminP}[1]{\text{argmin}}

\newcommand{\argmaxP}[1]{\text{argmax}}

\newcommand{\proofread}[1]{}
\newif\ifArxiv
\usepackage{xcolor}
\definecolor{Blue}{rgb}{0.2, 0.2, 0.8}
\definecolor{Black}{rgb}{0, 0, 0}
\usepackage{tikz}
\usetikzlibrary{matrix,calc}
\tikzstyle{block} = [draw,rectangle,thick,minimum height=2em,minimum width=2em]
\tikzstyle{arrow} = [->,thick]
\usepackage{pgfplots}
\pgfplotsset{width=10cm,compat=1.9,every axis plot/.append style={line width=2pt}} 
\makeatletter
\IEEEtriggercmd{\reset@font\normalfont\fontsize{7.0pt}{7.2pt}\selectfont}
\makeatother
\IEEEtriggeratref{1}

\makeatletter
\newcommand\fs@ruled@notop{\def\@fs@cfont{\bfseries}\let\@fs@capt\floatc@ruled
  \def\@fs@pre{}%
  \def\@fs@post{\kern2pt\hrule\relax}%
  \def\@fs@mid{\kern2pt\hrule\kern2pt}%
  \let\@fs@iftopcapt\iftrue}
\renewcommand\fst@algorithm{\fs@ruled@notop}
\makeatother

\title{\Large\bf Real-Time Decentralized Navigation of Nonholonomic Agents\\ Using Shifted Yielding Areas \vspace{-10px}
}
\author{Liang He$^{1}$, Zherong Pan$^{2}$, Dinesh Manocha$^{3}$  \\
\vspace{-30px}\\
\thanks{$^1$Liang He is with the University of North Carolina at Chapel Hill. \{lianghe.hust@gmail.com\} $^2$Zherong Pan is with the University of North Carolina at Chapel Hill. \{zherong.pan.usa@gmail.com\} $^3$Dinesh Manocha is with the University of Maryland, College Park. \{dm@cs.umd.edu\}}}
        
\begin{document}
\maketitle
\thispagestyle{empty}
\pagestyle{empty}

\begin{abstract}
We present a lightweight, decentralized algorithm for navigating multiple nonholonomic agents through challenging environments with narrow passages. Our key idea is to allow agents to yield to each other in large open areas instead of narrow passages, to increase the success rate of conventional decentralized algorithms. At pre-processing time, our method computes a medial axis for the freespace. A reference trajectory is then computed and projected onto the medial axis for each agent. During run time, when an agent senses other agents moving in the opposite direction, our algorithm uses the medial axis to estimate a Point of Impact (POI) as well as the available area around the POI. If the area around the POI is not large enough for yielding behaviors to be successful, we shift the POI to nearby large areas by modulating the agent's reference trajectory and traveling speed. We evaluate our method on a row of 4 environments with up to 15 robots, and we find our method incurs a marginal computational overhead of 10-30 ms on average, achieving real-time performance. Afterward, our planned reference trajectories can be tracked using local navigation algorithms to achieve up to a $100\%$ higher success rate over local navigation algorithms alone.
\end{abstract}
\section{Introduction}
In recent years, autonomous vehicles have been deployed in complex, city-scale scenarios to accomplish various tasks such as food delivery, warehouse administration, and public transportation. These vehicles routinely travel on highly regulated paths, such as highways, crossroads, and sidewalks, or in spaces with large open areas including shopping malls, school libraries, etc. Most prior works \cite{6225166,bareiss2015generalized,9196799,8715479} build navigation algorithms on one of these assumptions. In reality, however, autonomous vehicles must also be prepared for unexpected and unregulated scenarios or spaces with narrow passages. Dealing with narrow spaces is inevitable when two food delivery robot meets in the aisle of a hotel or an autonomous truck travels downtown to reach a warehouse. Narrow passages are notoriously difficult to handle, even when navigating a single robot \cite{szkandera2020narrow}, and scaling to multiple agents is still an open problem.

\begin{figure}[h]
\centering
\scalebox{.8}{\includegraphics[width=1\linewidth]{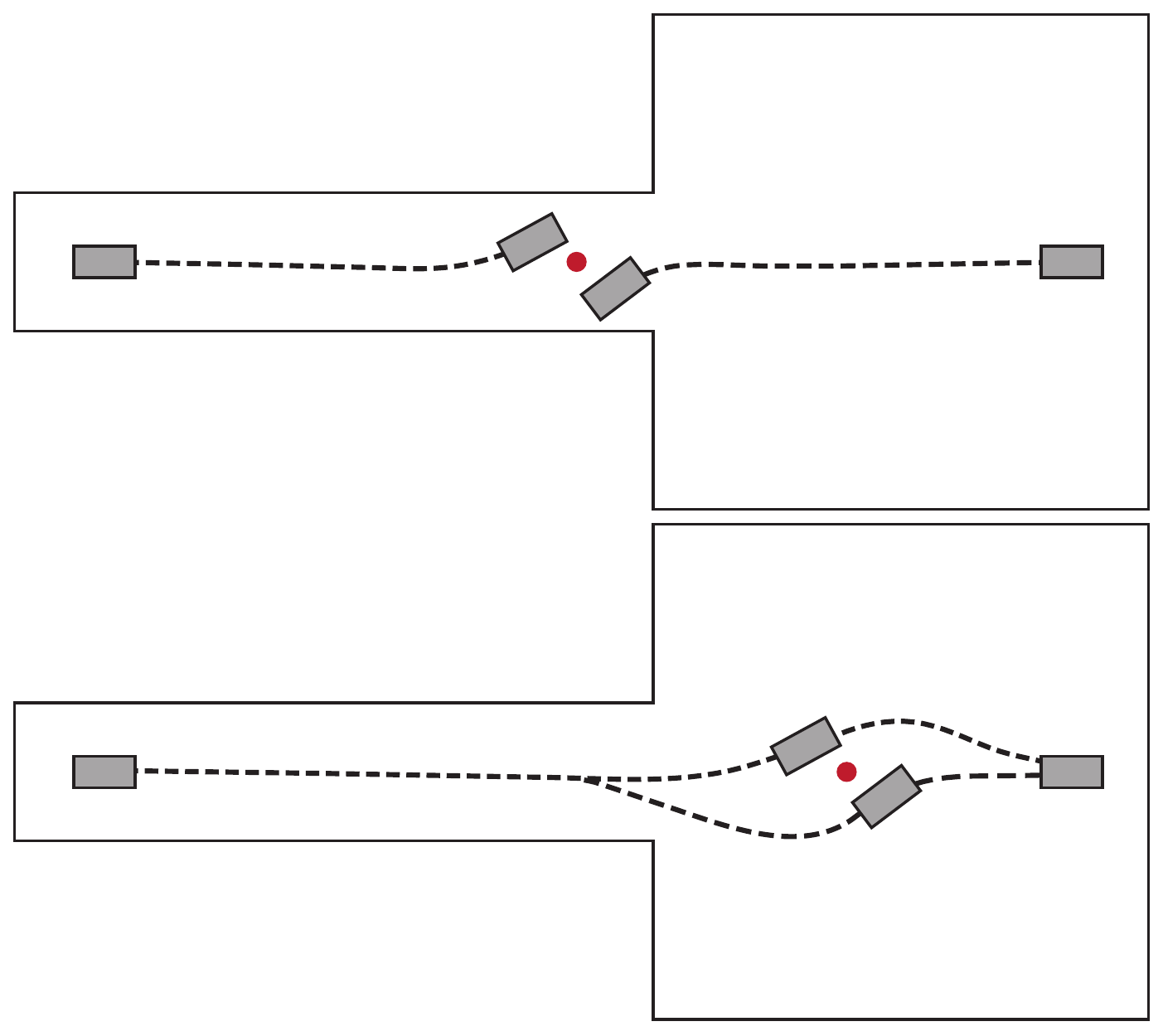}}
\caption{\label{fig:illus} \small{Two agents travel in an environment with one large and one narrow space. (a): Top picture: if the two agents travel at the same speed, their meeting point (defined as the POI and illustrated as the red dot) will be in the narrow space where local navigation techniques can fail. (b)Bottom picture: our method shifts the POI to the large space so that local navigation can successfully generate collision-free trajectories by yielding.}}
\vspace*{-10px}
\end{figure}
Prior methods for navigating multiple agents are classified into decentralized local techniques and centralized global techniques, each having its pros and cons. Local navigation methods \cite{6225166,bareiss2015generalized} assume agents move towards their goal positions along some local directions without communicating with each other. When obstacles or other agents get in the way, heuristic behaviors, such as yielding \cite{van2011reciprocal}, grouping\cite{6630970}, and following \cite{10.5555/2982818.2982838,7487147} are used to avoid collisions. However, local techniques can fail in the face of narrow passages where agents form deadlock configurations as illustrated in \prettyref{fig:illus} (a). On the other hand, global navigation methods \cite{Berg-RSS-09,yu2018effective,he2022multi} coordinate agent motions in a central node to avoid collisions. Although these methods can handle many agents in complex environments with narrow passages, they rely on strong assumptions such as the environment being grid-like, agents moving on discrete graph-like structures, or the agents being holonomic. However, actions such as constructing such discrete structures or generalizing to nonholonomic agents are non-trivial and cannot be used in time-critical applications due to a high computational cost.

\TE{Main Result:} We propose an improved decentralized algorithm for nonholonomic multi-agent navigation, which incorporates ideas from centralized techniques to alleviate the deadlock problem. We observe that yielding behaviors used by prior local navigation approaches \cite{6225166,bareiss2015generalized} can have high success rates in large open areas while being less successful in narrow spaces as illustrated in \prettyref{fig:illus} (b). As a result, we propose shifting the yielding areas to large open spaces of the environment to increase the success rate. Specifically, our algorithm relies on the construction of a medial axis for the free space. By mapping agent positions and their trajectories to the medial axis, we can estimate their Positions-Of-Impacts (POIs), which are positions where agents get close enough for local navigation techniques to generate yielding behaviors. We then estimate the surrounding space required by such yielding behaviors. If the space around a POI is not large enough for the yielding to be successful, we search for nearby large spaces and re-plan agent trajectories to move the POI. We show that such re-planning can be accomplished at a relatively low-cost without communication with other agents, preserving the decentralized nature of our method.

We evaluate our method in 4 challenging scenarios with 5-15 robots. The results show that our method exhibits real-time performance, taking up to 20 ms and 43 ms on average to plan the POIs. Compared with local navigation alone, our method achieves up to a $100\%$ higher success rate in some scenarios.
\section{Related Work}
Over the last two decades, a large body of works on the multi-agent narrow passage navigation problem in motion planning has emerged. 

Widely used sampling algorithms such as RRT~\cite{lavalle2001randomized} and PRM~\cite{kavraki1994probabilistic} can work in high-dimensional configuration spaces by, looking for feasible motion plans, and extensions including RRT$^*$~\cite{karaman2011anytime} and FMT$^*$~\cite{janson2015fast} can find (nearly) optimal trajectories. These algorithms have been extended to handle nonholonomic agents \cite{webb2013kinodynamic,doi:10.1177/0278364915614386}. Unfortunately, both theoretical analysis \cite{doi:10.1177/0278364917714338} and empirical studies \cite{szkandera2020narrow} have shown that such algorithms incur extremely high computational overheads. Indeed, narrow passages significantly reduce the set of the lookout \cite{619371}, which is crucial to the efficacy of sampling, while the complexity of optimal motion planning grows exponentially with the number of agents \cite{doi:10.1177/0278364917714338}. Almost all these algorithms are offline and inappropriate for time-critical applications such as autonomous driving.

Local navigation techniques use a set of heuristic rules to generate moving directions. These methods incur a much lower computational cost but sacrifice completeness or feasibility. In practice, however, they can have a high success rate under certain assumptions. Successful local navigation algorithms include the dynamic windows~\cite{fox1997dynamic}, reciprocal velocity obstacles (RVO)~\cite{van2011reciprocal,6225166,bareiss2015generalized}, and potential fields~\cite{koren1991potential,6943069}. All these methods were originally proposed for holonomic robots and extensions to differential drive models have been proposed. It is noteworthy that RVO and its variants can provide a collision-free guarantee, which allows agents to alter their moving directions or come to a full stop before collisions. This feature of RVO typically produces a yielding behavior allowing agents to move around local obstacles and continue towards the goal. However, the ambient space required for such yielding behaviors is generally larger for nonholonomic robots than holonomic ones, making RVO-based methods less successful in differential drive models and narrow passages. 

A different category of methods, known as centralized, global algorithms~\cite{yu2013multi,luna2011push,yu2018effective}, involves discretizing the agent motions on a grid or a graph-like structure. Graph search algorithms can then be used to find optimal~\cite{sharon2015conflict}, near optimal~\cite{yu2018effective}, or feasible trajectories~\cite{yu2015pebble} for large groups of agents within a relatively small computational budget. However, these methods are mostly designed for holonomic robots, and extensions to nonholonomic cases are far from trivial while their computational cost cannot meet real-time requirements. Our method can be interpreted as a special kind of centralized algorithm on the medial axis graph of the free space, on which we plan the POIs. The low-level yielding actions are then generated using local navigation techniques within each POI.

Finally, we have noticed some recent works~\cite{sun2017no,bareiss2015generalized,barrett2013teamwork,trautman2015robot} apply data-driven techniques to multi-agent navigation problems. By presenting agents with examples of optimal solutions in challenging scenarios, some learned policies can outperform analytic techniques. These techniques are parallel and orthogonal to our contribution. We speculate that learning-based techniques can be used as the local navigator in our method to generate high-quality yielding behaviors in large open areas. However, these methods incur a high computational cost in the training phase and re-training is required when the environment changes. The results of learned navigation policies are also sensitive to training parameters and network architectures. These potential drawbacks inspire us to design low-cost algorithms based on existing local navigation algorithms, with a higher success rate.

\section{Problem Formulation \& Background}
We assume there are $N$ nonholonomic agents with the configuration of $i$th agent being $x_i(t)$ at time instance the $t$. The agent moves in a 2D freespace $\mathcal{F}\subset\mathbb{R}^2$ according to the following differential drive model:
\begin{align*}
\dot{x}_i(t)=f_i(x_i,u_i),
\end{align*}
where $u_i$ is the control signal. With each agent starting from an initial configuration $x_i(0)$, our goal is to find $u_i(t)$ for $t\in[0,T]$ such that $p(x_i(T))$ is close enough to some goal position $g_i$, where $p(\bullet)$ is the configuration-to-position mapping function. Given $g_i$, local navigation algorithms~\cite{van2011reciprocal,bareiss2015generalized} would direct agents via a desired velocity $v_i^*$ and modulate $u_i$ to locally avoid collisions. We build our method on the generalized RVO algorithm denoted as a function:
\begin{align*}
u_i(t)\triangleq\text{GRVO}(v_i^*,x_i(t)).
\end{align*}
Such modulation typically exhibits yielding behaviors allowing a crowd of agents to move around each other and continue towards their respective goals. However, extra space is required for local yielding to be successful. This property is exploited in prior work~\cite{Solovey-RSS-15} to design centralized navigation algorithms for holonomic agents, while nonholonomic agents typically require even larger yielding space. The choice of desired velocity is another key to the success of local navigation. A prominent choice is $v_i^*\triangleq g_i-p(x_i(t))$, which is valid in open areas with small obstacles. For more complex or obstacle-rich environments, a set of reference trajectories must be computed to guide agents across large obstacles.

\begin{figure}[h]
\centering
\scalebox{.8}{\includegraphics[width=0.9\linewidth]{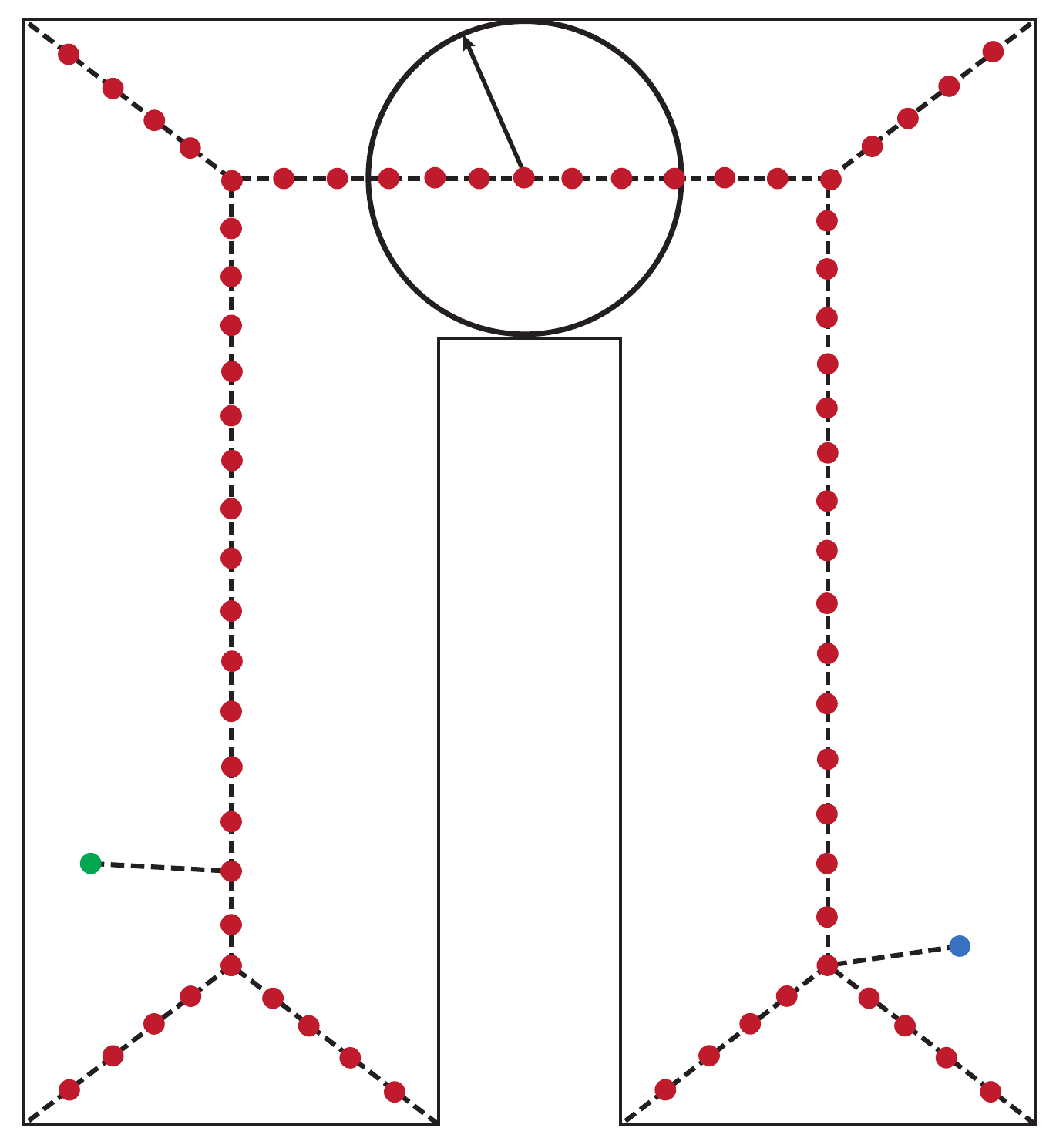}}
\caption{\label{fig:MedialAxis} \small{We illustrate the discretized medial axis for the U-shaped environment, where red dots belong to $V$ and black dashed edges belong to $E$. Each $s_i\in V$ is associated with a circular domain (black circle) with radius defined as $r(s_i)$ (black arrow). A reference trajectory is computed by first projecting $x_i(0)$ (blue) and $g_i$ (green) to $G$ and then computing the shortened path on $G$.}}
\vspace{-10px}
\end{figure}
\subsection{Blum Medial-Axis}
Our method makes extensive use of the medial axis of $\mathcal{F}$ to 1) estimate the area required by the yielding behavior and 2) compute reference trajectories. The definition of Blum medial-axis~\cite{blum1978shape} or skeleton is as follows. Given a 2D object defined by a \textbf{closed, oriented} boundary $\partial\mathcal{F}$, a Blum medial axis is a set. For every point $s$ in this set, we can find a unique circle centered at $s$ that is tangent to at least two points of $\partial\mathcal{F}$. This circle is known as the circular domain or domain of $s$ and we denote its radius as $r(s)$. A practical method like \cite{telea2006variational} would compute a discretized Blum medial axis, which is a graph $G=<V,E>$, where the set of vertices is sampled skeleton points $V=\{s_i\}$ at regular intervals connected by edges in $E$. As illustrated in \prettyref{fig:MedialAxis}, we compute a reference trajectory for the $i$th agent by first projecting $x_i(0)$ and $g_i$ to the closest vertices and then searching for a trajectory along $G$ via Dijkstra's algorithm.

\subsection{Trajectory Following with Yielding}
Given a reference trajectory, we have $x_i$ track the trajectory by designing the desired velocity $v_i^*$. Specifically, we set the desired velocity to be the negative gradient of a cost function $v_i^*\triangleq\nabla_{p(x_i)} -c(p(x_i))$ defined as:
\begin{align*}
c(p(x_i))\triangleq c_\text{follow}(p(x_i))+c_\text{bias}(p(x_i)),
\end{align*}
where $c_\text{follow}$ guides $x_i$ to move forward along the reference trajectory and $c_\text{bias}$ penalizes bias from the trajectory. We use an idea similar to the Frenet-frame-based tracking method \cite{werling2010optimal}. Specifically, we first compute the closest $s_i\in{V}$ to $x_i(t)$ that belongs to the reference trajectory. We denote $s_{i+1}$ as the next node in $V$ that also belongs to the reference trajectory, then we define:
\begin{align*}
c_\text{follow}(p(x_i))\triangleq& -(s_{i+1}-s_i)^T\dot{p}(x_i)\\
c_\text{bias}\triangleq& \|p(x_i)-s_i\|^2.
\end{align*}
In the next section, we describe a method to avoid deadlock configurations in narrow passages, allowing the yielding behaviors generated by $\text{GRVO}$ to have a high success rate.
\section{$\text{GRVO}$ with Shifted Yield Areas}
Our method differs from prior works by applying an additional modulation to the desired velocity function $v_i^*$ and we denote this function as $\mathcal{M}(v^*)$. The modulated velocity can be plugged into $\text{GRVO}$ to derive our final local navigation algorithm:
\begin{align*}
u_i(t)\triangleq\text{GRVO}(\mathcal{M}(v_i^*),x_i(t)).
\end{align*}
Note that our method can also be combined with local navigation methods other than $\text{GRVO}$. Our modulation function aims at shifting the POI between the two agents to large open areas in $\mathcal{F}$. Being a decentralized algorithm, such modulation is highly challenging because an agent does not have the ability to acquire other agents' trajectories, nor to alter their motions. However, we find it suffices to only modulate the velocity of the agent being considered based on a rough estimation of other agents' trajectories, as long as the same modulation function $\mathcal{M}$ is deployed on all the agents. In the following sections, we present details about POI detection, shifting, and modulation.

\subsection{POI Detection}
\begin{figure}[h]
\centering
\includegraphics[width=1\linewidth]{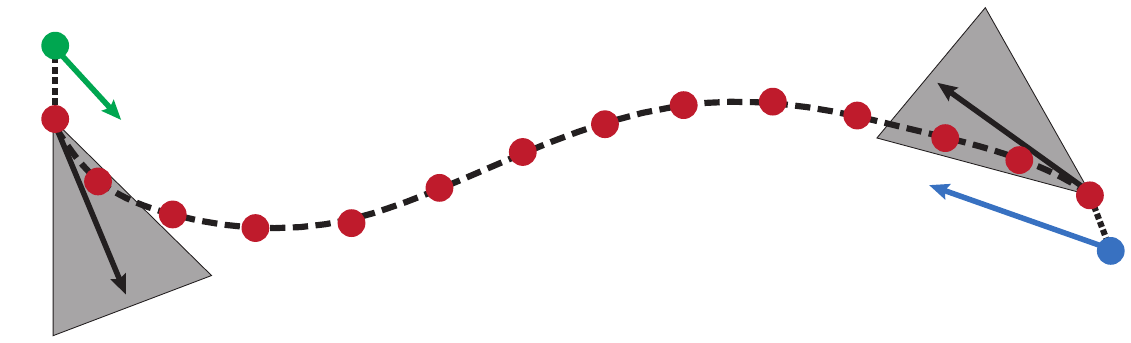}
\put(-225,65){$p(x_j)$}
\put(-25 ,5 ){$p(x_i)$}
\put(-247,50){$s_j$}
\put(-3  ,30){$s_i$}
\put(-215,45){$\dot{p}(x_j)$}
\put(-70 ,30){$\dot{p}(x_i)$}
\put(-205,5 ){$-\dot{P}_{ij}(1)$}
\put(-75 ,60){$\dot{P}_{ij}(0)$}
\put(-150,20){POI}
\caption{\label{fig:POI} \small{We illustrate the procedure of estimating the POI between $x_i$ (blue) and $x_j$ (green). Two points are first projected to the closest skeletal nodes $s_i$ and $s_j$, respectively. The shortened path $P_{ij}$ is illustrated with red nodes and dashed lines. If the difference between $\dot{p}(x_i)$, $\dot{p}(x_j)$ and tangents of $P_{ij}$ are smaller than a user-defined $\epsilon$ (gray cones) path, then we assume a POI exists. The POI is the point where two agents meet along $P_{ij}$. In this figure, since $x_j$ is slower (shorter green arrow), the POI is closer to $x_j$.}}
\vspace*{-10px}
\end{figure}
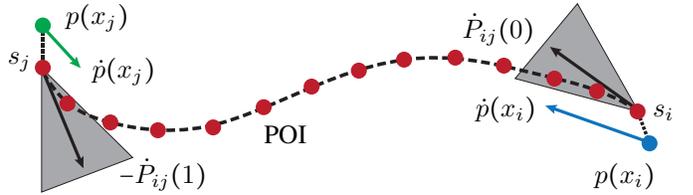
We define for each agent $x_i$ a sensing radius $R_i$. When any other agent $x_j$ satisfies $\|p(x_i)-p(x_j)\|\leq R_i$, we assume a potential yielding behavior might happen between them. Since $x_i$ does not know $x_j$'s future trajectory, we need to estimate POI based on the following assumption. We first project $p(x_i), p(x_j)$ onto their closest points on $G$, which  are denoted as $s_i$ and $s_j$, respectively. We then compute a shorted path between $s_i$ and $s_j$ on $G$ via Dijkstra's algorithm. In practice, we precompute the all-pair shortest distances so any shortened path can be looked up instantaneously. This path is denoted as $P_{ij}(\alpha)$, where $\alpha\in[0,1]$, $P_{ij}(0)=s_i$, and $P_{ij}(1)=s_j$. If both $x_i$ and $x_j$ are moving along the opposite tangential directions of $P_{ij}$, then we assume $P_{ij}$ is the estimated path containing a POI of the two agents. We determine that the two agents are traveling along opposite tangential directions if the following conditions hold:
\small
\begin{align}
\label{eq:meeting}
&\frac{\dot{P}_{ij}(0)}{\|\dot{P}_{ij}(0)\|}^T\frac{\dot{p}(x_i)}{\|\dot{p}(x_i)\|}>1-\epsilon \quad
-\frac{\dot{P}_{ij}(1)}{\|\dot{P}_{ij}(1)\|}^T\frac{\dot{p}(x_j)}{\|\dot{p}(x_j)\|}>1-\epsilon,
\end{align}
\normalsize
and no POI would be considered otherwise. Here $\epsilon$ is a user-defined upper bound of velocity bias. For a decentralized algorithm, our agent $x_i$ does not know the velocity of $x_j$ either, so we estimate $\dot{p}(x_j)$ using a finite difference of two consecutive frames of $x_j$. The POI between $x_i$ and $x_j$ is then estimated as $P_{ij}(\alpha_\text{POI})$ where $\alpha_\text{POI}$ is computed such that the following condition holds:
\begin{align*}
\frac{|P_{ij}([0,\alpha_\text{POI}])|}{\|\dot{p}(x_i)\|}=
\frac{|P_{ij}([\alpha_\text{POI},1])|}{\|\dot{p}(x_j)\|},
\end{align*}
where $|P_{ij}|$ denotes the arc-length of a sub-trajectory. The POI detection procedure is illustrated in \prettyref{fig:POI} and outlined in \prettyref{alg:POIDetection}, which incurs marginal overhead to conventional local navigation techniques.
\begin{algorithm}[t]
\caption{POISet($x_i$)\label{alg:POIDetection}}
\begin{algorithmic}[1]
\State Set$\gets\emptyset$
\For{Each agent $x_j\neq x_i\land\|p(x_j)-p(x_i)\|<R$}
\State Project to skeletal point $s_i,s_j$
\State Loop up shortest path $P_{ij}$
\If{\prettyref{eq:meeting} holds}
\State Set$\gets\text{Set}\bigcup\{P_{ij}(\alpha_\text{POI})\}$
\EndIf
\EndFor
\State Return Set
\end{algorithmic}
\end{algorithm}

\subsection{POI Shifting\label{sec:shifting}}
Given a POI located at $s_i\in V$, we then estimate its surrounding area. Given the medial axis, this area can be immediately estimated as the circular domain at $s_i$. If a POI is located on an edge of $E$ neighboring $s_i$ and $s_j$, we interpolate the circular domain radius. The radius of circular domain $r(s_i)$ must be sufficiently large for the yielding behavior to have a high success rate. Unfortunately, we are still lacking a theoretical analysis connecting the success rate of $\text{GRVO}$ and the size of the yielding area. Instead, we use the following heuristic rule to compute the minimal domain radius $r(s_i)$ for $n$ agents to successfully yield to each other:
\begin{align}
\label{eq:heuristic}
r(s_i)\geq \eta r(n+1),
\end{align}
where $\eta\in(0,1]$ is a user-provided parameter. In typical scenarios, we have $n=2$ since POI is estimated for two agents. If \prettyref{eq:heuristic} is violated, we need to shift POI to a nearby large space on the medial axis graph $G$. We propose first searching for nodes belonging to $P_{ij}$. This is because $P_{ij}$ lies on our estimated path and shifting POI within $P_{ij}$ would not cause a detour. If $P_{ij}$ does not contain any node satisfying \prettyref{eq:heuristic}, we search the entire $G$ for the nearest node, which is the center of a large domain. If both attempts fail, we decide the entire map consists of narrow spaces and do not shift POI. This procedure is summarized in \prettyref{alg:POIShifting}.
\begin{algorithm}[t]
\caption{POIShift(POI,$n$)\label{alg:POIShifting}}
\begin{algorithmic}[1]
\State Dist$\gets\infty$, $\text{POI}_0\gets\text{POI}$, POI$\gets$None
\For{Each $s_i\in P_{ij}$}
\If{\prettyref{eq:heuristic}$\land\|\text{POI}_0-p(s_i)\|<\text{Dist}$}
\State $\text{POI}\gets p(s_i)$, Dist$\gets\|\text{POI}_0-p(s_i)\|$
\EndIf
\EndFor
\If{Dist$<\infty$}
\State Return $\text{POI}$
\EndIf
\For{Each $s_i\in V$}
\If{\prettyref{eq:heuristic}$\land\|\text{POI}_0-p(s_i)\|<\text{Dist}$}
\State $\text{POI}\gets p(s_i)$, Dist$\gets\|\text{POI}_0-p(s_i)\|$
\EndIf
\EndFor
\State Return $\text{POI}$
\end{algorithmic}
\end{algorithm}

\subsection{POI Merging}
We found that handling only POI cases with two agents improves the success rate of $\text{GRVO}$. For extremely challenging environments, however, more agents can meet at nearby POIs and we must consider POIs involving $n>2$ agents. We handle this case by iteratively merging nearby POIs as outlined in \prettyref{alg:POIMerging}. In practice, if two POIs denoted as $\text{POI}_i$ and $\text{POI}_j$ involve $n_i$ and $n_j$ agents, respectively, we merge them into a single $\text{POI}_{ij}$ if the following condition holds:
\begin{align}
\label{eq:mergeCondition}
\|\text{POI}_i-\text{POI}_j\|\leq\eta\min\left(r(n_1+1),r(n_2+1)\right).
\end{align}
The merged $\text{POI}_{ij}$ involves $n=n_i+n_j$ agents and its required yielding radius is specified by \prettyref{eq:heuristic}. We perform the POI shifting procedure as described in \prettyref{sec:shifting}. If the shifting procedure fails, then we reject merging. We iteratively merge POIs until no more merging can be performed.
\begin{algorithm}[t]
\caption{POIMerge($x_i$)\label{alg:POIMerging}}
\begin{algorithmic}[1]
\State Set$\gets$POISet($x_i$), More$\gets$True
\For{$\text{POI}\in$Set}
\If{POIShift(POI,2)$\neq$None}
\State{Set$\gets$Set$/\{\text{POI}\}$}
\State{Set$\gets$Set$\bigcup$POIShift(POI,2)}
\EndIf
\EndFor
\While{More}
\State More$\gets$False
\For{A pair of $\text{POI}_i, \text{POI}_j\in$Set with $n_i, n_j$}
\If{\prettyref{eq:mergeCondition} holds}
\If{POIShift($\text{POI}_i$,$n_i+n_j$)$\neq$None}
\State{Set$\gets$Set$/\{\text{POI}_i,\text{POI}_j\}$}
\State{Set$\gets$Set$\bigcup$POIShift($\text{POI}_i$,$n_i+n_j$)}
\State More$\gets$True
\ElsIf{POIShift($\text{POI}_j$,$n_i+n_j$)$\neq$None}
\State{Set$\gets$Set$/\{\text{POI}_i,\text{POI}_j\}$}
\State{Set$\gets$Set$\bigcup$POIShift($\text{POI}_j$,$n_i+n_j$)}
\State More$\gets$True
\EndIf
\EndIf
\EndFor
\EndWhile
\State Return Set
\end{algorithmic}
\end{algorithm}

\begin{figure*}[t]
\centering
\scalebox{.85}{
\setlength{\tabcolsep}{3px}
\begin{tabular}{cccc}
\includegraphics[width=0.24\linewidth,trim=2px 2px 2px 2px,clip]{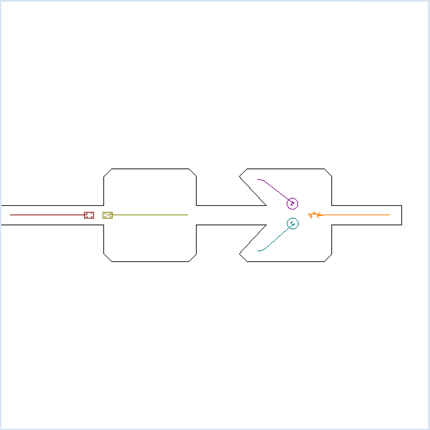}\put(-90,5){(a)}\put(-110,90){GRVO+following}&
\includegraphics[width=0.24\linewidth,trim=1px 1px 1px 1px,clip]{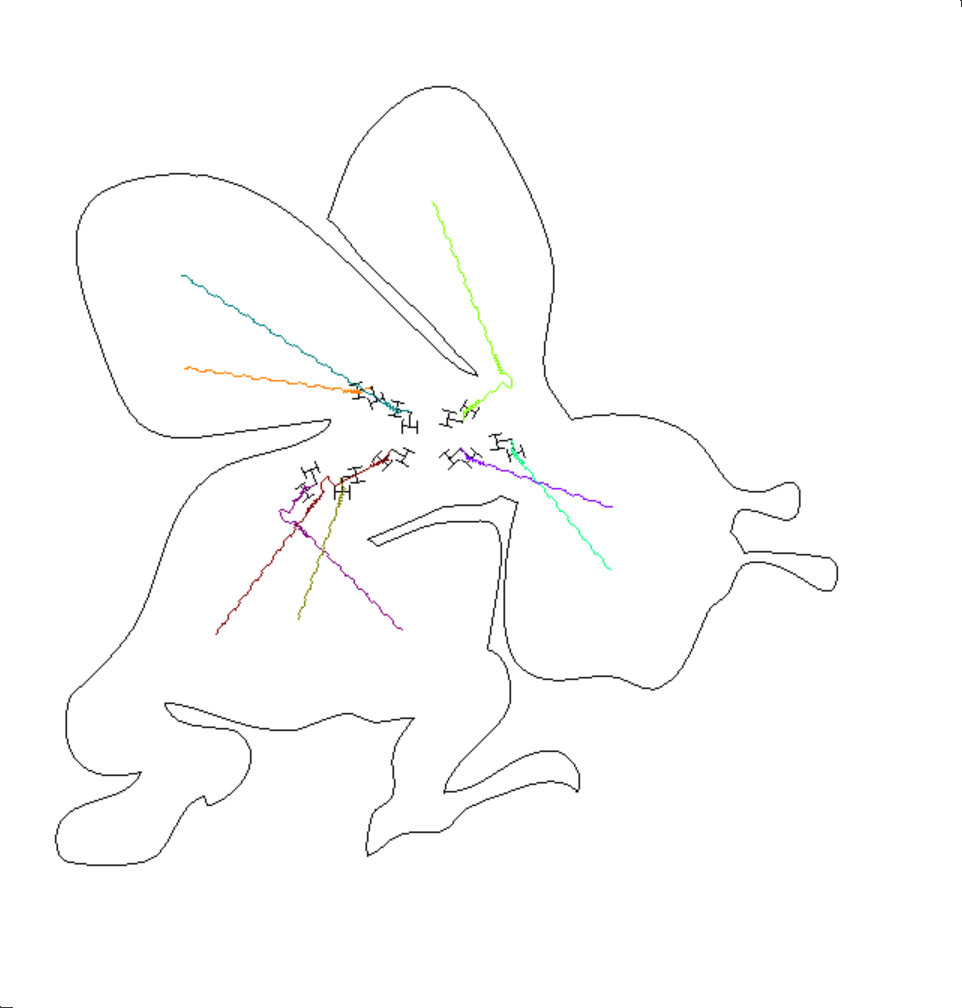}\put(-90,5){(b)}&
\includegraphics[width=0.24\linewidth,trim=6px 6px 6px 6px,clip]{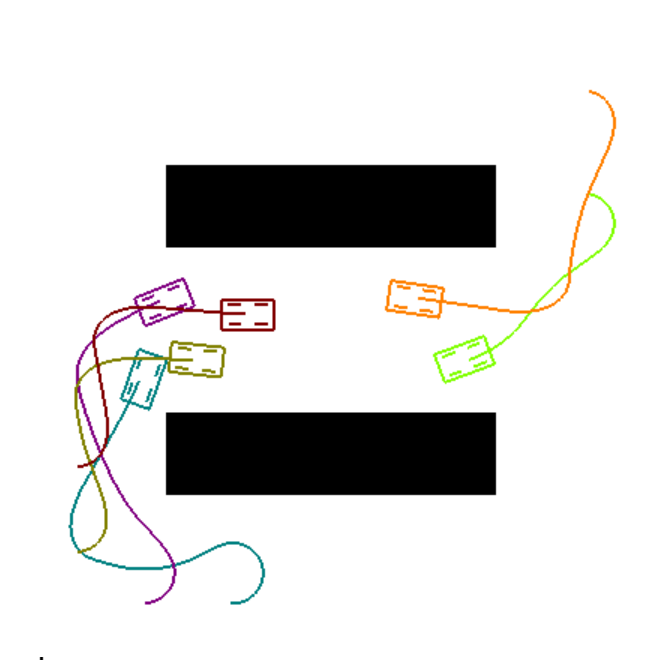}\put(-90,5){(c)}&
\includegraphics[width=0.24\linewidth,trim=6px 6px 6px 6px,clip]{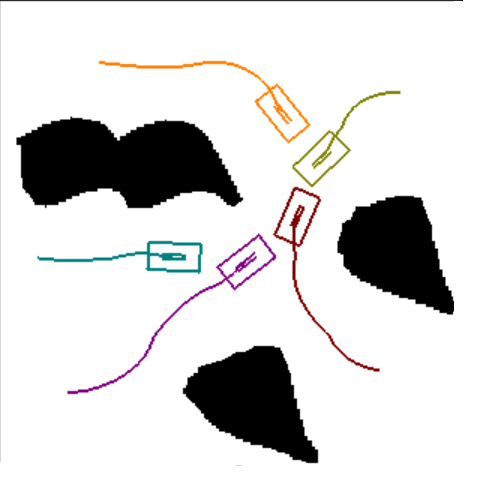}\put(-90,5){(d)}\\
\includegraphics[width=0.24\linewidth,trim=2px 2px 2px 2px,clip]{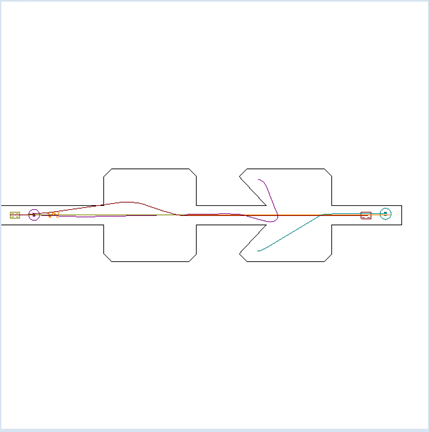}\put(-110,90){Bridge}&
\includegraphics[width=0.24\linewidth,trim=1px 1px 1px 1px,clip]{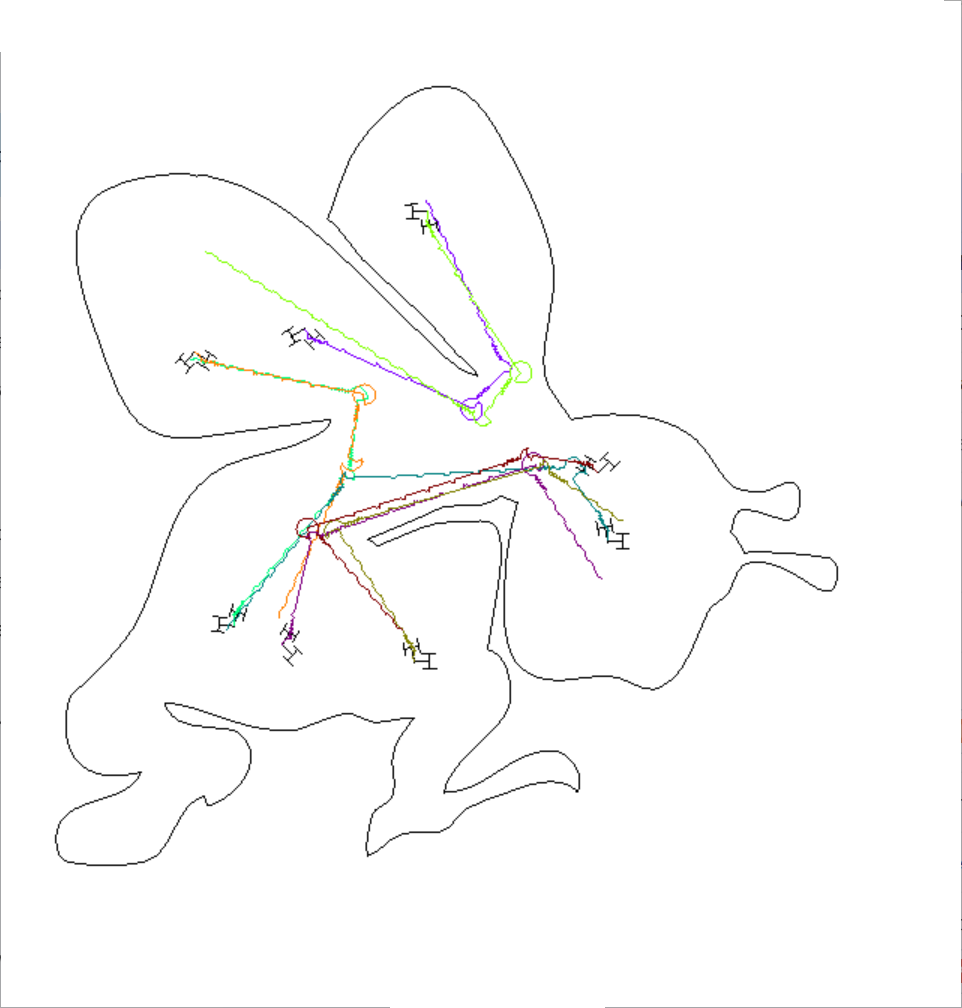}&
\includegraphics[width=0.24\linewidth,trim=6px 6px 6px 6px,clip]{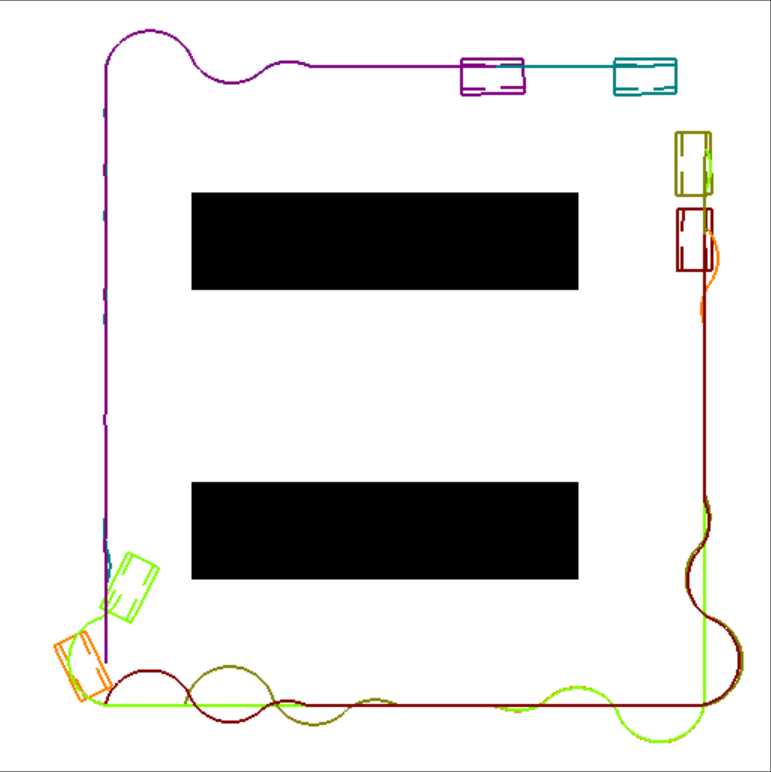}&
\includegraphics[width=0.24\linewidth,trim=6px 6px 6px 6px,clip]{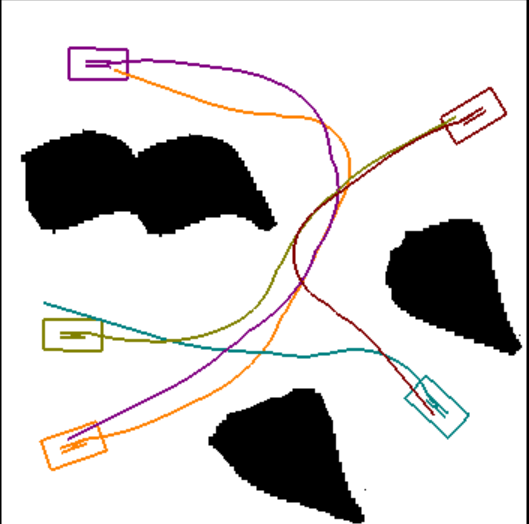}\\
\includegraphics[width=0.24\linewidth,trim=2px 2px 2px 2px,clip]{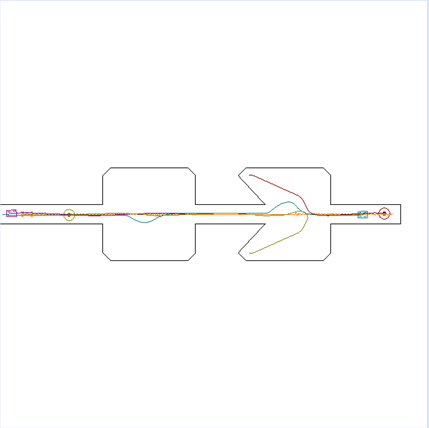}\put(-110,90){Ours}&
\includegraphics[width=0.24\linewidth,trim=1px 1px 1px 1px,clip]{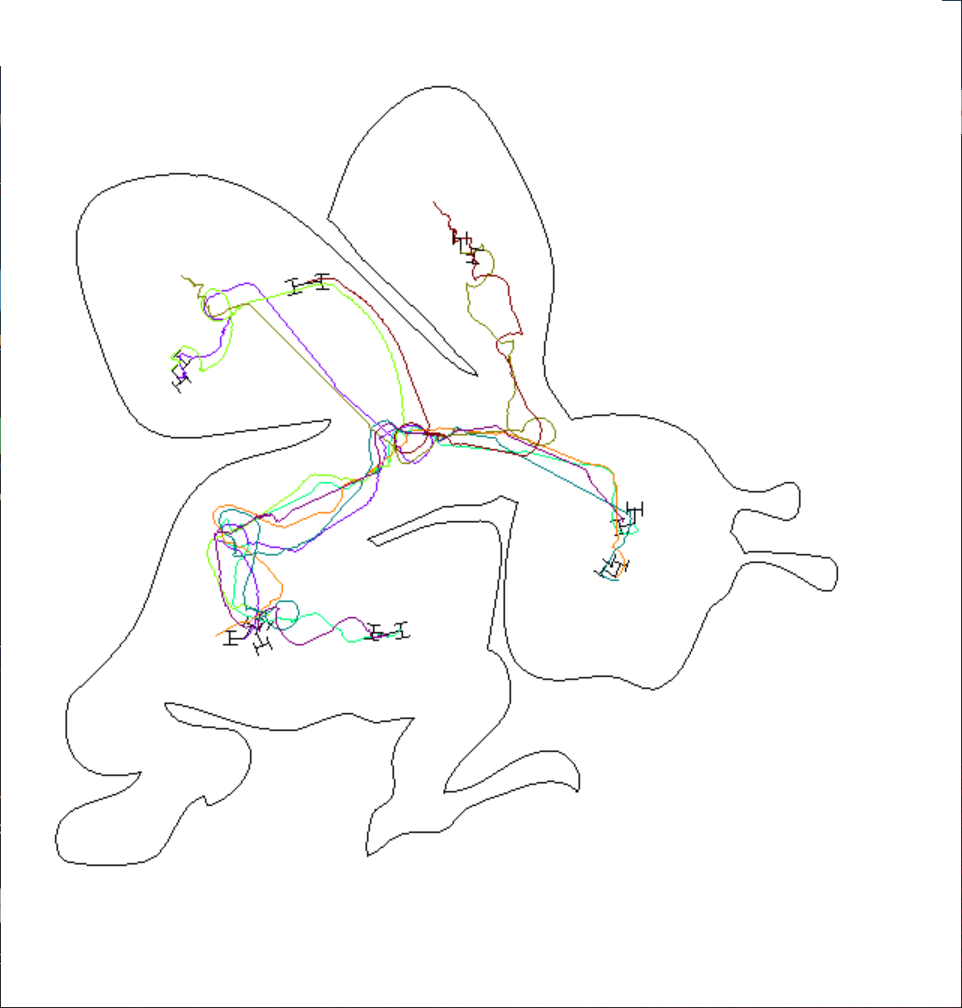}&
\includegraphics[width=0.24\linewidth,trim=6px 6px 6px 6px,clip]{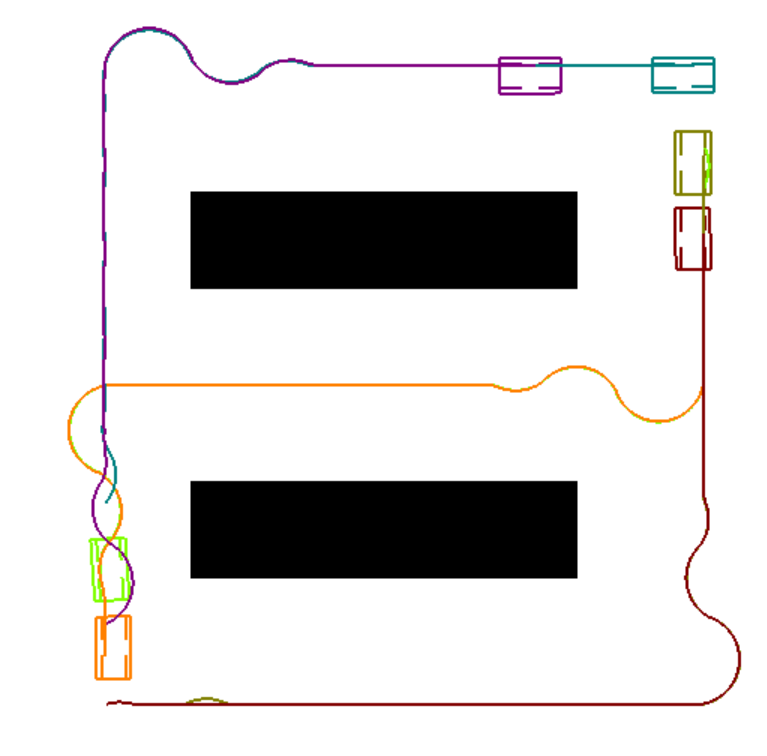}&
\includegraphics[width=0.24\linewidth,trim=6px 6px 6px 6px,clip]{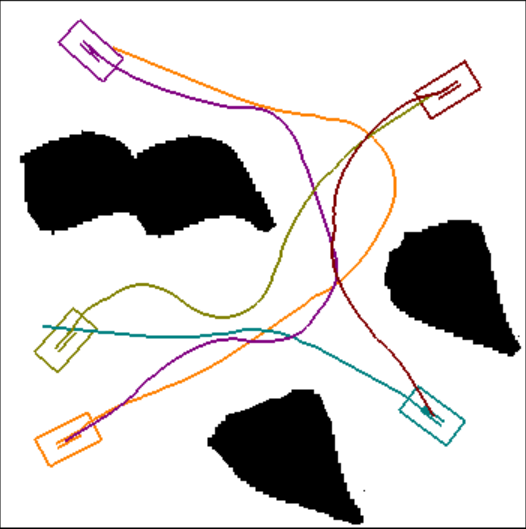}\\
\end{tabular}}
\caption{\label{fig:results} \small{The 4 most challenging benchmarks used in our experiments with agent trajectories computed using GRVO+following, Bridge, and our method are highlighted in the first, second, and third rows, respectively: (a): We random generate several agents in two open areas connected by a narrow passage; (b): Random agents are generated in a complex Bee-shaped environment; (c): A maze with two long obstacles creating narrow aisles in between; (d): A narrow passage with a garage in the middle. Agents traveling through the narrow passage must move to the garage temporarily to accomplish yielding behaviors.}}
\end{figure*}

\begin{table*}[t]
\centering
\resizebox{\textwidth}{!}{
\begin{tabular}{|c|c|c|c|c|c|c|c|c|c|c|c|c|}
\hline
\multirow{2}{*}{Method} & 
\multicolumn{3}{c|}{Benchmark 1} & 
\multicolumn{3}{c|}{Benchmark 2} & 
\multicolumn{3}{c|}{Benchmark 3} &  
\multicolumn{3}{c|}{Benchmark 4}\\
 & Traj. Length & Succ. Rate & FPS & 
   Traj. Length & Succ. Rate & FPS & 
   Traj. Length & Succ. Rate & FPS & 
   Traj. Length & Succ. Rate & FPS\\
\hline \hline
GRVO     & 1311 & 0\%   & 45 & 632 & 4\%   & 43 & 433 & 50\%  & 55 & $401$ & 50\% & $60$\\
GRVO+IC  & 1258 & 0\%   & 35 & 562 & 12\%  & 33 & 341 & 94\%  & 42 & $333$ & 100\% & 55\\
Bridge   & 607  & 100\% & - & 424 & 100\% & - & 324 & 100\% & - & $312$ & 100\% & $-$\\
Ours     & 805  & 100\% & 41 & 455 & 100\% & 43 & 421 & 100\% & 42 & $387$ & 100\% & 43\\
\hline
\end{tabular}}
\caption{\label{table:profile} \small{We compare our approach with previous methods, GRVO, GRVO+following, and Bridge, on the 4 benchmarks, in terms of average agent trajectory length (we only consider trajectories of agents that successfully reach goals), success rate over $50$ random scenarios (we only consider a scenario successfully handled when all the agents reach their goals), and the FPS.}}
\vspace{-15px}
\end{table*}
\subsection{Velocity Modulation}
After the above procedure, an agent has a set of POI positions against a multitude of other agents. We choose the nearest $\text{POI}$ to $p(x_i(t))$ as the temporary goal point to modulate our velocity. Note that due to various sources of uncertainty and inaccuracy in estimating $\text{POI}$, modulating our velocity can cause detours. To minimize this effect, we only adopt modulation if the nearest $\text{POI}$ was successfully shifted, i.e. we define $\mathcal{M}$ as:
\begin{align*}
\mathcal{M}(v_i^*)\triangleq
\begin{cases}
\left[\argmin{\text{POI}}\|\text{POI}-p(x_i)\|\right]-x_i & \text{POI shifted}   \\
v_i^* & \text{otherwise}.
\end{cases}
\end{align*}

\subsection{Acceleration by Precomputation}
The main computational bottleneck of our algorithm lies in the POI merging procedure. This involves at most $2N$ calls to \prettyref{alg:POIShifting}, and each call to \prettyref{alg:POIShifting} incurs a computational cost of $|V|+|E|$. However, we can further reduce the cost of \prettyref{alg:POIShifting} to $\mathcal{O}(1)$ by precomputing a lookup table. Note that POI generally lies on an edge of $G$. However, if we use sufficiently dense samples to construct $V$, we can shift POI to a nearby vertex $s_i\in V$ incurring a small error. In this way, \prettyref{alg:POIShifting} will only be called with discrete inputs POIShift($s_i$,$n$), and we can construct a table of size $|V|\times N$ to precompute all possible results. After such acceleration, the complexity of each evaluation of modulation function $\mathcal{M}$ is only $\mathcal{O}(2N)$.
\section{Evaluation}
\begin{figure}[ht]
\centering
\includegraphics[width=0.7\linewidth]{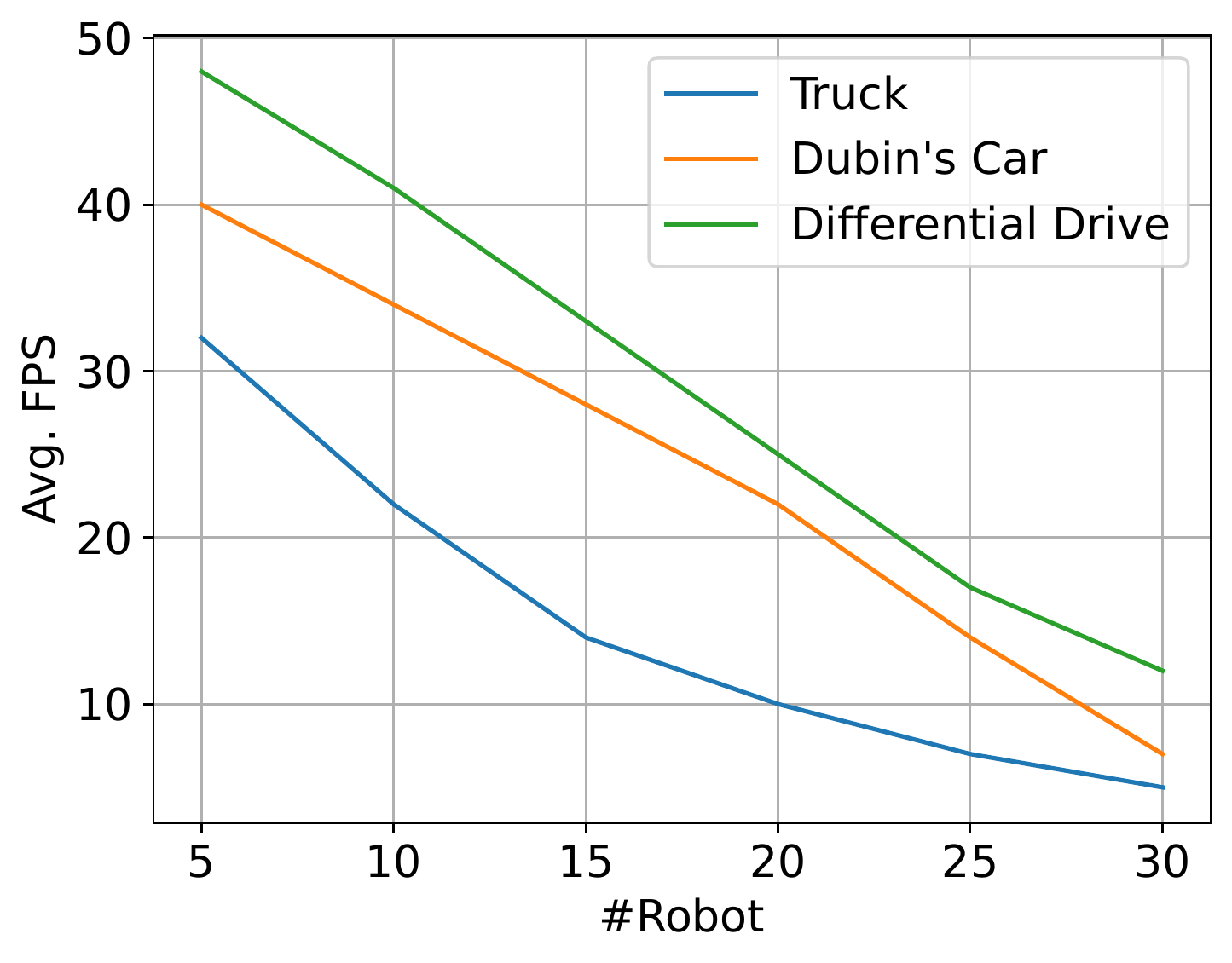}
\caption{\label{fig:FPS} \small{We plot the averaged computational cost of our method (frames per second) against the number of robots.}}
\vspace{-15px}
\end{figure}

We have implemented our algorithms in C++ on an Intel Core i7 CPU running with 16GB of RAM. We use the CGAL library to build the medial axis graph $G$. We evaluate our method on three categories of robots: the single differential-drive robot, the Dubin's car, and the differential-drive robot with trailer (truck for short) as in \cite{bareiss2015generalized}, where we tune the parameters such that the maximal turning curvature of the trajectory is $0.19$ for a Dubin's car, and $0.22$ for a truck-like robot. In all three testing scenarios, we use GRVO \cite{bareiss2015generalized} as our local navigation algorithm. We set the vehicle size to $4\times5$ square units and, the medial axis sampling interval to $0.02$ units, and we use $\eta=1.6-2.6, R=15-30$. We randomly put the agents in the open area and repeat 50 times in each scenario. The average computational cost over these scenarios is summarized in~\prettyref{fig:FPS}, and it largely depends on the number of robots. Our bottleneck lies in the collision detection between robots of non-circular shapes.

\subsection{Baselines}
\begin{wraptable}{r}{0.25\textwidth}
\centering
\scalebox{.8}{
\begin{tabular}{ccc}
\toprule
    & Bridge  &  Ours   \\
\midrule
I   & 212     & 0.6     \\
II  & 122     & 3.3     \\
III & 12      & 0.2     \\
IV  & 19      & 4       \\
\bottomrule
\end{tabular}}
\caption{\label{table:precompute}\small{This table shows the average computational time (seconds) during the precomputation stage of Bridge and our method on 4 scenarios.}}
\end{wraptable}
There are several prior works on improving the success rate of local navigation methods. Our first baseline is the GRVO algorithm~\cite{bareiss2015generalized} without our modulation. Our second baseline is the grouped sampling-based algorithm (Bridge)~\cite{he2017efficient}. This algorithm aims at solving the same problem as ours. They use a sampling-based method to precompute a set of corridors across narrow passages, in which nonholonomic agent trajectories can be efficiently generated by interpolation. Agents follow these interpolated trajectories in the corridor while collisions are handled using local navigation techniques. Finally, we also consider the GRVO algorithm with adapted Implicit Coordination method (GRVO+following)~\cite{7487147}. As the major difference from our method, the IC algorithm allows an agent to communicate with neighbors to coordinate desired velocities.

\subsection{Benchmark Problems}
We consider 4 challenging benchmarks, illustrated in \prettyref{fig:results}. The trajectories generated by different methods are compared and evaluated in three aspects: the average length of agent trajectories, the rate of success of finding feasible motion plans, and the frame rate per second (FPS). Quantitative results, corresponding to an average of over $50$ simulations with randomly generated agent configurations in an assigned sub-area of the scenarios, are summarized in \prettyref{table:profile}.

\TE{Benchmark I:} We use a dumb-like environment, shown in \prettyref{fig:results} (a), where multiple agents move from one side to the other. We observe that our method allows the agents to determine that they will meet agents from the other side and the POIs lie in the narrow central passage. Our method then has agents on one side retreat from the narrow space and return to the left side of the scene to wait for agents from the other side to pass through before moving on. For this example, most other local navigation methods, including GRVO and GRVO+following, fail. 

\TE{Benchmark II:} As shown in \prettyref{fig:results} (b), we place a group of agents in a complex, bee-shaped environment. Agents start from one corner of the freespace and repeatedly yield other upcoming agents. The results in \prettyref{table:profile} show that our method can always compute a feasible motion plan, while prior techniques cause many collisions resulting in deadlock configurations. The only rival algorithm that exhibits a high success rate is Bridge, which uses a sampling-based motion planner during the precomputation stage. In comparison, the precomputation involved in our method is only used to find the medial axis graph of $\mathcal{F}$, which can be accomplished much faster as profiled in \prettyref{table:precompute}.

\TE{Benchmark III:} As shown in \prettyref{fig:results} (c), we use a small maze involving two long obstacles, with agents again placed randomly. The Bridge algorithm outperforms our method for this benchmark in terms of trajectory length, although the success rates of both algorithms are $100\%$. This is due to inaccuracies in detecting and shifting POIs, where our method does not account for in-between obstacles.

\TE{Benchmark IV:} Our last and most challenging benchmark involves a single narrow passage with a garage in the middle for agents to perform yielding. As illustrated in \prettyref{fig:results} (d), our POI shifting procedure allows agent to be directed to the garage, while all prior methods fail.

\section{Conclusion \& Limitation}
We propose a novel velocity modulation algorithm to improve the success rate of prior local navigation algorithms for multiple nonholonomic agents. We observe that local navigation methods can generate yielding behaviors for agents so they can move around each other and continue toward their respective goals. However, yielding requires extra space and can have a low success rate in narrow passages. To alleviate this problem, we propose shifting the POI between two or more agents to large open areas. We show that even using a rough estimation of POI and the required space for yielding, such a strategy can empirically improve the success rate of conventional local navigation algorithms such as GRVO~\cite{bareiss2015generalized} by 100\% in some scenarios. A major issue with the current method is our decentralized setting, which does not allow any communication or coordination between agents. It is thus difficult to further improve the accuracy of POI estimation and shifting. In further works, we are considering extending our method to allow local communications between agents to achieve partial coordination, as has been done in \cite{hildreth2019coordinating}.

\AtNextBibliography{\footnotesize}
\printbibliography
\end{document}